\documentclass{article}

\usepackage{microtype}
\usepackage{graphicx}
\usepackage{booktabs} %

\usepackage{hyperref}

\usepackage[accepted]{icml2024}

\usepackage{amsmath,amsfonts,bm}

\def\eqref#1{equation~\ref{#1}}

\def\1{\bm{1}}

\DeclareMathAlphabet{\mathsfit}{\encodingdefault}{\sfdefault}{m}{sl}
\SetMathAlphabet{\mathsfit}{bold}{\encodingdefault}{\sfdefault}{bx}{n}

\newcommand{\softmax}{\mathrm{softmax}}

\DeclareMathOperator*{\argmax}{arg\,max}
\DeclareMathOperator*{\argmin}{arg\,min}

\usepackage{hyperref}
\usepackage{url}
\usepackage{graphicx}
\usepackage{booktabs}
\usepackage{siunitx}
\usepackage{adjustbox}

\usepackage[utf8]{inputenc} %
\usepackage[T1]{fontenc}    %
\usepackage{hyperref}       %
\usepackage{url}            %
\usepackage{booktabs}       %
\usepackage{amsfonts}       %
\usepackage{nicefrac}       %
\usepackage{microtype}      %
\usepackage{xcolor}         %

\usepackage{dcolumn}
\newcolumntype{d}[1]{D{.}{.}{#1}}

\usepackage{amsthm}
\usepackage{amsmath}
\usepackage{amssymb}
\usepackage{breqn}
\usepackage{dsfont}
\usepackage{mathtools} 
\usepackage{subfig}
\usepackage{xspace}

\usepackage{array}
\newcolumntype{L}[1]{>{\raggedright\let\newline\\\arraybackslash\hspace{0pt}}m{#1}}
\newcolumntype{C}[1]{>{\centering\let\newline\\\arraybackslash\hspace{0pt}}m{#1}}
\newcolumntype{R}[1]{>{\raggedleft\let\newline\\\arraybackslash\hspace{0pt}}m{#1}}

\usepackage{wasysym}

\usepackage{algorithm}
\usepackage{algpseudocode}

\DeclareMathOperator*{\topk}{topk}
\DeclareMathOperator*{\lowerbound}{lowerbound}
\DeclareMathOperator*{\upperbound}{upperbound}

\DeclareMathOperator*{\clip}{clip}

\newcommand{\norm}[1]{\left\lVert#1\right\rVert}

\newcommand{\etal}[0]{et al.\xspace} %

\newcommand{\eat}[1]{}

\DeclareFontFamily{U}{mathx}{}
\DeclareFontShape{U}{mathx}{m}{n}{<-> mathx10}{}
\DeclareSymbolFont{mathx}{U}{mathx}{m}{n}
\DeclareMathAccent{\widehat}{0}{mathx}{"70}
\DeclareMathAccent{\widecheck}{0}{mathx}{"71}

\renewcommand{\checkmark}{\CIRCLE\xspace}
\newcommand{\emptymark}{\Circle\xspace}
\newcommand{\halfmark}{\LEFTcircle\xspace}

\icmltitlerunning{Et Tu Certifications: Robustness Certificates Yield Better Adversarial Examples}%

\begin{document}

\twocolumn[
\icmltitle{Et Tu Certifications: Robustness Certificates Yield Better Adversarial Examples}

\begin{icmlauthorlist}
\icmlauthor{Andrew C. Cullen}{yyy}
\icmlauthor{Shijie Liu}{yyy}
\icmlauthor{Paul Montague}{comp}
\icmlauthor{Sarah M. Erfani}{yyy}
\icmlauthor{Benjamin I.P. Rubinstein}{yyy}
\end{icmlauthorlist}

\icmlaffiliation{yyy}{School of Computing and Information Systems, University of Melbourne, Parkville, Australia}
\icmlaffiliation{comp}{Defence Science and Technology Group, Adelaide, Australia}

\icmlcorrespondingauthor{Andrew C. Cullen}{andrew.cullen@unimelb.edu.au}

\icmlkeywords{Machine Learning, ICML}

\vskip 0.3in
]

\printAffiliationsAndNotice{}%

\begin{abstract}

In guaranteeing the absence of adversarial examples in an instance's neighbourhood, certification mechanisms play an important role in demonstrating neural net robustness. In this paper, we ask if these certifications can compromise the very models they help to protect? Our new \emph{Certification Aware Attack} exploits certifications to produce computationally efficient norm-minimising adversarial examples $74 \%$ more often than comparable attacks, while reducing the median perturbation norm by more than $10\%$. While these attacks can be used to assess the tightness of certification bounds, they also highlight that releasing certifications can paradoxically reduce security.
\end{abstract}

\section{Introduction}\label{sec:introduction}

A troubling property of learned models is that semantically indistinguishable samples can trigger different model outputs~\citep{biggio2013evasion}. Known as \emph{adversarial examples} when constructed deliberately, these samples can be incredibly difficult to detect, especially when the distance to clean examples is minimised. While \emph{adversarial defences} have been proposed as a best-response countermeasure, the security they provide is often illusory as they can be exploited or evaded by motivated attackers.

As a response to this dynamic, guarantees that no adversarial examples exist within a calculable, bounded region---through techniques known as Certified Robustness---have emerged~\citep{weng2018towards,zhang2018efficient,li2018certified,salman2019convex, cullen2022double}.

However while these certifications are typically framed as providing security, they cannot intrinsically distinguish between clean or adversarial examples. This is a consequence of \emph{adversarial examples also being able to be certified}, and that practical attacks still exist \emph{outside} the region of certification. While the existence of norm-minimising adversarial examples that satisfy this condition should not be controversial, we strongly argue that the certification community has been underestimating the impact of how these techniques are framed on non-expert users. Presenting certification techniques as being robust to adversarial examples has the potential to induce a false sense of security in users who are not familiar with the inherent risks facing these models. 

\begin{figure*}[!ht]
    \centering
    \includegraphics[width=.65\linewidth]{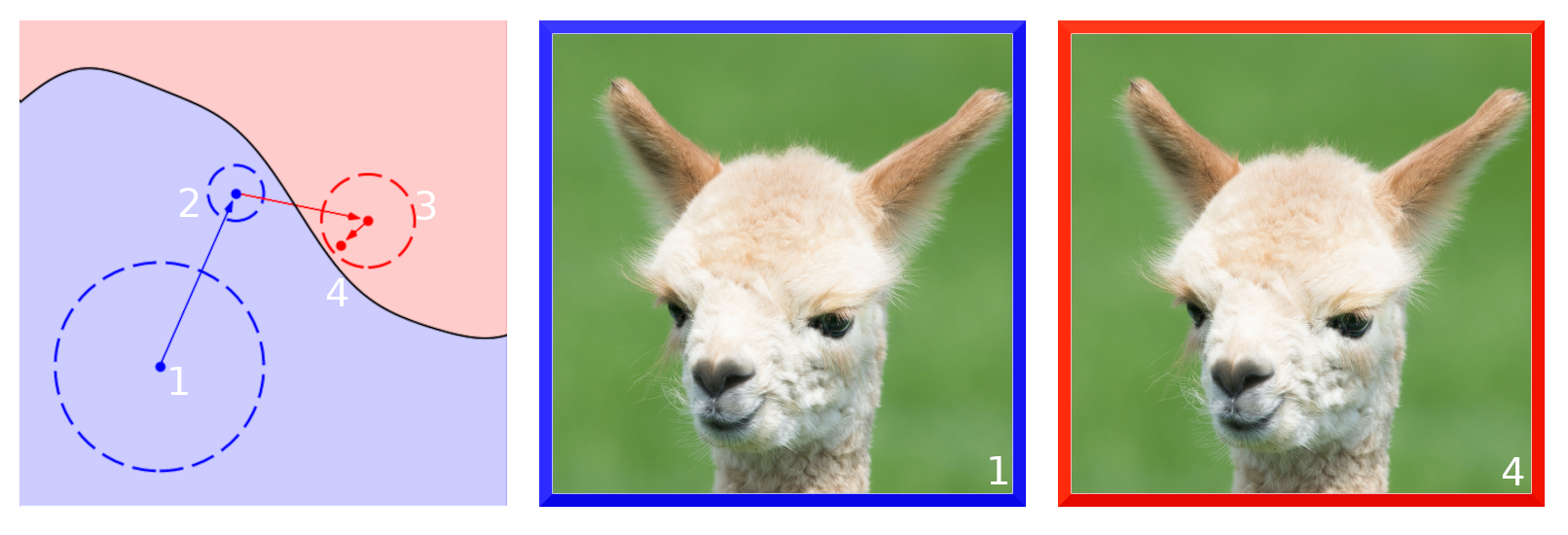} 
  \caption{Illustrative example of an evasion attack for a binary classifier, that changes the output from blue to red. Our new attack framework exploits knowledge of the certifications (circles) to minimise the number of iterative steps required.  }\label{fig:llama_progression}   
\end{figure*}

Within this work we demonstrate that this risk is amplified by the fact that \emph{certifications can be exploited to construct smaller adversarial perturbations} than prior approaches. This attack, which we will henceforth refer to as a \emph{Certification Aware Attack} exploits the very nature of certifications to assist in rapidly identifying adversarial examples---a process that is demonstrated in Figure~\ref{fig:llama_progression}. Exploiting certifications allows our new attack framework to \emph{(i) speed up the initial stages of the search with larger and more informative jumps, and (ii) to reduce the total adversarial perturbation}, which ensures that these adversarial examples have a higher chance of avoiding detection~\citep{gilmer2018motivating}. As part of this, we also introduce a cohesive framework for attacking certified models.

The utility of such an attack extends to improving our understanding of the tightness of calculated certifications---for if a certification is a guaranteed lower bound on the location of the nearest adversarial example, a norm-minimising attack corresponds to an upper bound of the same quantity. However, while there is beneficial utility in these attacks, their existence still demonstrates that certification researchers must fundamentally reconsider how we think about the security implications of certifications. We believe that the only appropriate defence is to treat the certifications and class expectations as highly confidential information that must not be released, lest it be used to help compromise the very models the certification mechanisms purport to defend.

\section{Background: Certification Mechanisms}\label{sec:certification_mechanisms}

Certification mechanisms eschew the responsive view of adversarial defences in favour of bounding the region in which adversarial examples $\mathbf{x}'$ can exist for a given input sample $\mathbf{x}$---typically $B_p(\mathbf{x}, r)$ a $\mathbf{x}$-centred $p$-norm ball of some radius $r$. 
To be \emph{sound}, a radius must be strictly less than
\begin{align}\label{eqn:r_definition}
    r^\star &= \inf\left\{ \|\mathbf{x}-\mathbf{x}'\|_p : \mathbf{x}'\in \mathcal{S}, F(\mathbf{x})\neq F(\mathbf{x}') \right\} \\
    &\text{           where }
    F(\cdot) = \mathds{1} \left(\argmax_{i \in \mathcal{K}} f_i(\cdot)\right) \enspace.\nonumber
\end{align}
Here $\mathds{1}$ is a one-hot encoding of the predicted class in $\mathcal{K} = \{1, \ldots, K\}$, and $\mathcal{S}$ is the permissible input space, for example $[0,1]^d$ is typical in computer vision. The size of $B_p(\mathbf{x}, r)$ can be considered a reliable proxy for both the \emph{detectability} of adversarial examples~\citep{gilmer2018motivating} and the \emph{cost} to the attacker~\citep{huang2011adversarial}. 

These certificates can be constructed through either exact or statistical-sampling based methods. Exact methods typically construct their bounds by way of either interval bound propagation (IBP), which propagates interval bounds through the model; or convex relaxation, which utilises linear relaxation to construct bounding output polytopes over input bounded perturbations~\cite{salman2019convex, mirman2018differentiable, weng2018towards, zhang2018efficient, singh2019abstract, mohapatra2020towards}, in a manner that generally provides tighter bounds than IBP~\cite{lyu2021towards}. However, these techniques exhibit a time or memory complexity that makes them infeasible for complex model architectures or high-dimensional data~\cite{wang2021beta, chiang2020certified, levine2020randomized}. While Lipschitz certified mechanisms~\cite{tsuzuku2018lipschitz, leino2021globally} have recently been proposed as a less computationally intensive alternative than bound propagation mechanisms, they still exhibit scalability issues for larger and more semantically complex models. 

Statistical-sampling based methods build upon %
\emph{randomised smoothing}~\citep{lecuyer2019certified}, in which the Monte Carlo estimator of the expectation  under repeatedly perturbed sampling 
\begin{eqnarray}\label{eqn:expectations}
     \frac{1}{N} \sum_{j=1}^{N} F(\mathbf{X}_j) &\approx& \mathbb{E}_{\mathbf{X}}[F(\mathbf{X})]  \qquad \forall i \in \mathcal{K} \\
        \mathbf{X}_1, \ldots, \mathbf{X}_N, \mathbf{X} &\stackrel{i.i.d.}{\sim}& \mathbf{x} + \mathcal{N}(0, \sigma^2)\enspace, \nonumber
\end{eqnarray}
can be exploited to provide guarantees of invariance under \emph{additive} perturbations. In forming this aggregated classification, the model is re-construed as a \emph{smoothed classifier}, which in turn is certified. Approaches for constructing such certifications include differential privacy~\citep{lecuyer2019certified,dwork2006calibrating}, R\'{e}nyi divergence \citep{li2018certified}, and parameterising worst-case behaviours \citep{cohen2019certified, salman2019provably, cullen2022double}. The latter has proved the most performant, and yields certifications of
\begin{equation}\label{eqn:Cohen_Bound}
r = \frac{\sigma}{2} \left( \Phi^{-1}\left(\widecheck{E}_{0}[\mathbf{x}]\right) - \Phi^{-1}\left(\widehat{E}_{1}[\mathbf{x}]\right) \right)\enspace.
\end{equation}
Here $\Phi^{-1}$ is the inverse standard normal CDF, 
$(E_0, E_1) = \topk\left(\left\{\mathbb{E}_{\mathbf{X}}[F(\mathbf{X})]\right\}, 2\right)$
, and $(\widecheck{E}_0, \widehat{E}_1)$ are the lower and upper confidence bounds of these quantities to some confidence level $\alpha$ \citep{goodman1965simultaneous}. 

\section{Attacking Certified Defences}\label{sec:attacking_randomised}

That certifications are framed as being robust to adversarial manipulation has troubling security implications, as it may induce a false sense of security in those unfamiliar with these techniques limitations.
This is especially true when as a certification cannot tell us if a sample has already been compromised, and does not obviate the existence of semantically indistinguishable adversarial examples. However, as Table~\ref{tab:attack_properties} shows, to date only \citet{cohen2019certified} has consider test-time attack against certified models. However, they explicitly note that their tested attack does not align with the concept of attacking a certified model, and present no quantitative measures of performance. While other works have considered certifications from an adversarial lens, their focused has been on manipulating the training corpus to improve test time performance---a process that does not require reliable, hard to detect norm-minimising attacks. Thus there currently is clearly an insufficient understanding of both the tightness of certification bounds, and the risks facing certified models.

\begin{table*}
  \caption{Extant attacks, distinguished by if their goal is to change the label of samples or to just improve robust accuracy; if they were deployed at train- or test-time; if the attack has direct applicability to certifications (where half-circles denote attempting to improve certified robustness); and if they exploit the certifications themselves.
  }
  \label{tab:attack_properties}
  \centering
  \resizebox{1.3\columnwidth}{!}{%
  \begin{tabular}{lcC{1.1cm}C{1.1cm}C{1.15cm}C{1.25cm}}
    \toprule
        & & \multicolumn{3}{c}{\textbf{Applicability}} & \\
        \cline{3-5}
    \textbf{Algorithms} & \textbf{Goal} & \textbf{Train} & \textbf{Test} & \textbf{Certified} & \textbf{Exploits Certs.} \\ %
    \midrule
    PGD \citep{madry2017towards} & Label & \checkmark & \checkmark & \emptymark & \emptymark \\ %
    \cite{carlini2017towards} & Label & \checkmark & \checkmark & \emptymark & \emptymark  \\ %
    AutoAttack \citep{croce2020reliable} & Label & \checkmark & \checkmark & \emptymark & \emptymark  \\ %
    DeepFool \cite{moosavi2016deepfool} & Label & \emptymark & \checkmark & \emptymark & \emptymark  \\ %
    Training w/ Noise \citep{bishop1995training}  & Acc. & \checkmark & \emptymark & \halfmark &  \emptymark \\
    \cite{salman2019provably} & Acc. & \checkmark & \emptymark &  \halfmark & \emptymark  \\ %
    MACER \citep{zhai2020macer} & Acc. & \checkmark & \emptymark & \halfmark & \emptymark \\ 
    \citet{cohen2019certified} & Label & \emptymark & \checkmark & \checkmark & \emptymark \\
    \midrule
    Ours & Label & \halfmark & \checkmark & \checkmark & \checkmark  \\ %
    \bottomrule
  \end{tabular}
  }
\end{table*}

When it comes to attacking certified models, one may intuitively think of simply identifying a sample $\mathbf{x}'$ such that $\|\mathbf{x}' - \mathbf{x}\| = r^{\star}$. However, in practice certification mechanisms are not able to construct tight bounds on $r^{\star}$, and even if they were, the search space for identifying $\mathbf{x}'$ would still be significant. And even if such a point could be identified, its certified radii would be $0$, which would likely trigger further inspection in any operationalised certification system. As such, within this work, we introduce the idea of a \emph{confident} adversarial attack against a certification mechanism being one in which a certification constructed at the adversarial example is non-zero. For randomised smoothing this condition is equivalent to%
\begin{align} \label{eqn:general_robust}
\argmax \mathbb{E}_{\mathbf{X}}\left[F(\mathbf{x}') \right] & \neq  \argmax \mathbb{E}_{\mathbf{X}}\left[ F(\mathbf{x}) \right] \;\; \mbox{and} \\
\widecheck{E}_{0}\left[F(\mathbf{x}')\right] &> \widehat{E}_{1}\left[ F(\mathbf{x}')\right]\enspace. \nonumber
\end{align}
That these expectations are highly concentrated (for sufficiently high Monte Carlo sample sizes) enables any of the reference attacks within Table~\ref{tab:attack_properties} to be effectively employed \emph{against the class expectations, rather than the individual draws under noise}. This contrasts with approaches like Expectation Over Transformation~\citep{athalye2018synthesizing}, in which each sample under noise is attacked in a numerically inefficient manner.%

A complicating factor is that in randomised smoothing is that the final model layers can either be defined differentiable $\softmax$ or non-differentiable $\argmax$ layers~\citep{cullen2024-SP}. Most certification mechanisms assume the layer, including that of Equation~\ref{eqn:Cohen_Bound}. While it could naively be assumed that non-differentiable layers inherently defeat the gradient based attack mechanisms, in practice interventions like stochastic gradient estimation~\citep{fu2006gradient, chen2019fast}, surrogate modelling, and transfer attacks all providing potent mechanisms for a motivated attacker.

While we could test the performance of adversarial attacks under each of these interventions, distinguishing the impact of the attack against that of the intervention would be problematic. To both minimise the impact of ensuring differentiability and maximise the difficulty to the attacker, within this work we assume that the final $\argmax$ layer can be replaced with a Gumbel Softmax~\citep{jang2016categorical} %
\begin{equation}\label{eqn:gumbel}
    y_i = \frac{\exp\left((\log(\pi_i) + g_i) / \tau \right)}{\sum_{j \in \mathcal{K}} \exp\left((\log(\pi_i) + g_i) / \tau \right)}\;,\;\; \text{ } \forall i \in \mathcal{K}\enspace.
\end{equation}

We emphasise that this re-parametrisation is not necessary for the following attacks to work, as they can be applied to models with $\softmax$ outputs, or by way of any of the previously discussed interventions.

\subsection{Threat model}

Within this work we consider an attacker that attempts to construct a norm-minimising, confident (see Equation~\ref{eqn:general_robust}) adversarial example against a certified model. When the certification has been constructed through randomised smoothing, we assume the attacker has the ability to construct derivatives through $\argmax$ layers, and knowledge of the level of additive noise $\sigma$. However, we note that even this last assumption is not strictly necessary, as Appendix~\ref{app:sigma_accuracy} demonstrates that approximate values of $\sigma$ still provide enhanced certification performance. When attacking models that do not incorporate randomised smoothing, the attacker is assumed to have white-box access to the models gradients, predictions, and certifications.

\section{Certification Aware Attacks}\label{sec:CAA}

While there is value in understanding the tightness of certification mechanisms by attacking them with extant attacks, within this work we are also interested in understanding how certifications may be exploited by a motivated attacker to minimise the size of the identified examples. Such a concept may seem contradictory, but it is important to consider that from an attacker's perspective a certification can be viewed as a \emph{lower bound on the space where attacks may exist.} 

Section~\ref{sec:specifying} demonstrates how the existence of certifications at all points across the instance space~\citep{cullen2022double} can be exploited to significantly reduce the search space for identifying adversarial examples. Once an adversarial example is identified, Section~\ref{sec:refining} then demonstrates how certifications associated with successful adversarial examples can be \emph{exploited to minimise the perturbation norm of the sample}, as any norm-minimising step inside the certified radii still remains an adversarial example!

\subsection{Step Size Control}\label{sec:specifying}

We begin our attack by introducing the surrogate problem
\begin{equation}\label{eqn:surrogate_problem}
\hat{\mathbf{x}} =  \argmin_{\hat{\mathbf{x}}\in\mathcal{S}} \left\{\left|E_0(\hat{\mathbf{x}}) - E_1(\hat{\mathbf{x}})\right| \; : \;  F(\hat{\mathbf{x}}) = F(\mathbf{x}) \right\}\enspace.
\end{equation}
This formalism may seem counter-intuitive, as the constraint ensures that $\hat{\mathbf{x}}$ cannot be an adversarial example. However, consider the gradient-based solution of the previous problem
\begin{equation}\label{eqn:CAA_iter}
    \mathbf{x}_{i+1} = P_{\mathcal{S}}\left(\mathbf{x}_{i} - \epsilon_i \left(\frac{\nabla_{\boldsymbol{x}_i} | E_0[\mathbf{x}_i] - E_1[\mathbf{x}_i] |}{\norm{ \nabla_{\boldsymbol{x}_i} | E_0[\mathbf{x}_i] - E_1[\mathbf{x}_i] |}} \right)\right)\enspace,
\end{equation}
for a projection to the feasible space $P_{\mathcal{S}}$, and for which each $\mathbf{x}_i$ has associated certifications $r_i$. By imposing that $\epsilon_i > r_i$, we ensure that the new candidate solution $\mathbf{x}_{i+1}$ must exist outside the region of certification of the previous point, which is a \emph{necessary but not sufficient} condition for identifying an adversarial example.

Ensuring that  $\epsilon_i > r_i$ could be achieved by imposing that
\begin{equation}\label{eqn:epsilon_basic}
    \epsilon_i = \rho(\mathbf{x}_i) \left(1 + \delta\right)\enspace,
\end{equation}
for some $\delta > 0$, and where $\rho(\mathbf{x}_i) = r_i$. However, doing so fails to account for the information gained from the certifications at all $\mathbf{x}_{j}$ for $j = 0, \ldots, i$. If we instead construct 
\begin{align}\label{eqn:epsilon_complex}  
\rho(\mathbf{x}_i) &= \inf \left\{\hat{\rho}\geq 0 : \mathbf{x}^{\star}(\hat{\rho}) \notin  \bigcup_{j=0}^{i} B_P(\mathbf{x}_{j}, H\left[c_{0} = c_{j}\right] r_j)\right\}\\ %
\mathbf{x}^{\star}(\hat{\rho}) &= P_{\mathcal{S}}\left(\mathbf{x}_{i} - \hat{\rho} \left(\frac{\nabla_{\boldsymbol{x}_i} | E_0[\mathbf{x}_i] - E_1[\mathbf{x}_i] |}{\norm{ \nabla_{\boldsymbol{x}_i} | E_0[\mathbf{x}_i] - E_1[\mathbf{x}_i] |}} \right)\right)\enspace, \nonumber  
\end{align}
then $\mathbf{x}_{i+1}$ remains strictly outside the certified radii region of all $\{\mathbf{x}_i | c_i = c_0\}$. Here $c_i$ is the class prediction at step $i$ of the iterative process, and $H_{c_{0} = c_{i}}$ is an indicator function , and $\rho(\mathbf{x}_i)$ is solved for using a binary search. %

As taking large steps may be disadvantageous in certain contexts, in practice we define $\epsilon_i$ in terms of pre-defined lower- and upper-bounds
\begin{equation}\label{eqn:step_size_limits}
\tilde{\epsilon}_i = \clip\left(\epsilon_i, \epsilon_{\text{min}}, \epsilon_{\text{max}} \right)\enspace.
\end{equation}

\subsection{Refining Adversarial Examples}\label{sec:refining}

Once we have identified an adversarial example, we switch to the second stage of our iterative process, in which we minimise the perturbation norm of any identified examples, in order to decrease their detectability. At this stage, the attack iterates $\mathbf{x}_i$ now produces a class prediction of $c_i \neq c_0$. Thus, any $\mathbf{x}_i$ must also be an adversarial attack if the difference between the two points is less than or equal to $r_i$. Thus our iterator can be defined as
\begin{align}\label{eqn:correction}
\mathbf{x}_{i+1} &= P\left(\mathbf{x}_{i} - \min\{  \rho, \epsilon_{\text{max}} \} (1 - \delta) \left( \frac{\mathbf{x}_0 - \mathbf{x}_i}{\|\mathbf{x}_0 - \mathbf{x}_i\|}  \right)\right) \nonumber \\
\rho &= \sup \left\{ \hat{\rho}\geq 0 : \mathbf{x}^{\star}(\hat{\rho}) \in  \bigcup_{j=0}^{i} B_P(\mathbf{x}_{j}, H\left[c_{0} \neq c_{j}\right] r_j)\right\} \\
\mathbf{x}^{\star}(\hat{r}) &= P_{\mathcal{S}}\left(\mathbf{x}_{i} - \hat{\rho} \left( \frac{\mathbf{x}_0 - \mathbf{x}_i}{\|\mathbf{x}_0 - \mathbf{x}_i\|}  \right)\right)
\enspace. \nonumber    
\end{align}
Similar to Section~\ref{sec:specifying}, a simpler variant of the above simply involves setting that $\rho = r_i$, however doing so discards the potential for prior certifications to help refine the search space. This framing ensures that $c_{i} = c_{j} \text{ } \forall \text{ } j > i$---i.e., that all adversarial examples share the same class as the first identified adversarial example. As such, it may be true that there exists some adversarial example $\mathbf{x}''$
$$\|\mathbf{x}'' - \mathbf{x}_0\| < \|\mathbf{x}_{i} - \mathbf{x}_0\| \qquad \forall \text{ } i \in \mathbb{N} \enspace.$$
However, even with this limitation, the following section demonstrates that this process still produces significantly smaller adversarial examples than other techniques.

Algorithms detailing the aforementioned processes can be found within Appendices~\ref{app:algorithms} and \ref{sec:parameter_space}, and the code associated with this work can be found at \url{https://github.com/andrew-cullen/Attacking-Certified-Robustness}.%

\section{Results}\label{sec:results}

To demonstrate the performance of our new Certification Aware Attack, we test our attack relative to a range of other comparable approaches. We emphasise that both our new attack and the reference attacks are  \emph{deployed against certified models}, rather than the associated base classifiers.%

To achieve this, our experiments consider attacks against MNIST \citep{lecun1998gradient} (GNU v3.0 license), CIFAR-$10$ \citep{krizhevsky2009learning} (MIT license), and the Large Scale Visual Recognition Challenge variant of ImageNet \citep{deng2009imagenet, russakovsky2015imagenet} (which uses a custom, non-commercial license). In the case of models defended by randomised smoothing, each model was trained in PyTorch~\citep{NEURIPS2019_9015} using a ResNet-$18$ architecture, with experiments considering two distinct levels of smoothing noise scale $\sigma$. Additional experiments on the MACER~\citep{zhai2020macer} certification framework and a ResNet-$110$ architecture can be found in Appendix~\ref{app:macer}. The confidence intervals of expectations in all experiments were set according to the $\alpha = 0.005$ significance level. To demonstrate the generality of our identified threat model, Section~\ref{sec:attacking_ibp} eschews randomised smoothing to attack certifications constructed using IBP. Due to the inherent computational cost associated with constructing solutions with IBP, our results were limited to MNIST models solved using a sequential model of two convolutional layers followed by two linear layers, with ReLU activation functions. All calculations were constructed using one NVIDIA A100 GPUs for MNIST and CIFAR-$10$, while Imagenet test and training time evaluations employed two.

To explore the performance of our attacks, we will rely upon a handful of key metrics, including some that we have constructed specifically for this work. These include the success rate of successful attacks; the proportion of samples for which an attack produces smaller radii perturbations than its competitors; the percentage increase \emph{above} the certified radius $\%$-C; the median $\ell_2$ attack size $r_{50}$; and the attack time, which includes the time for certifying each sample. Further details can be found in Appendix~\ref{sec:metrics}

\begin{figure*}
\begin{center}
    \includegraphics[width=0.65\textwidth]{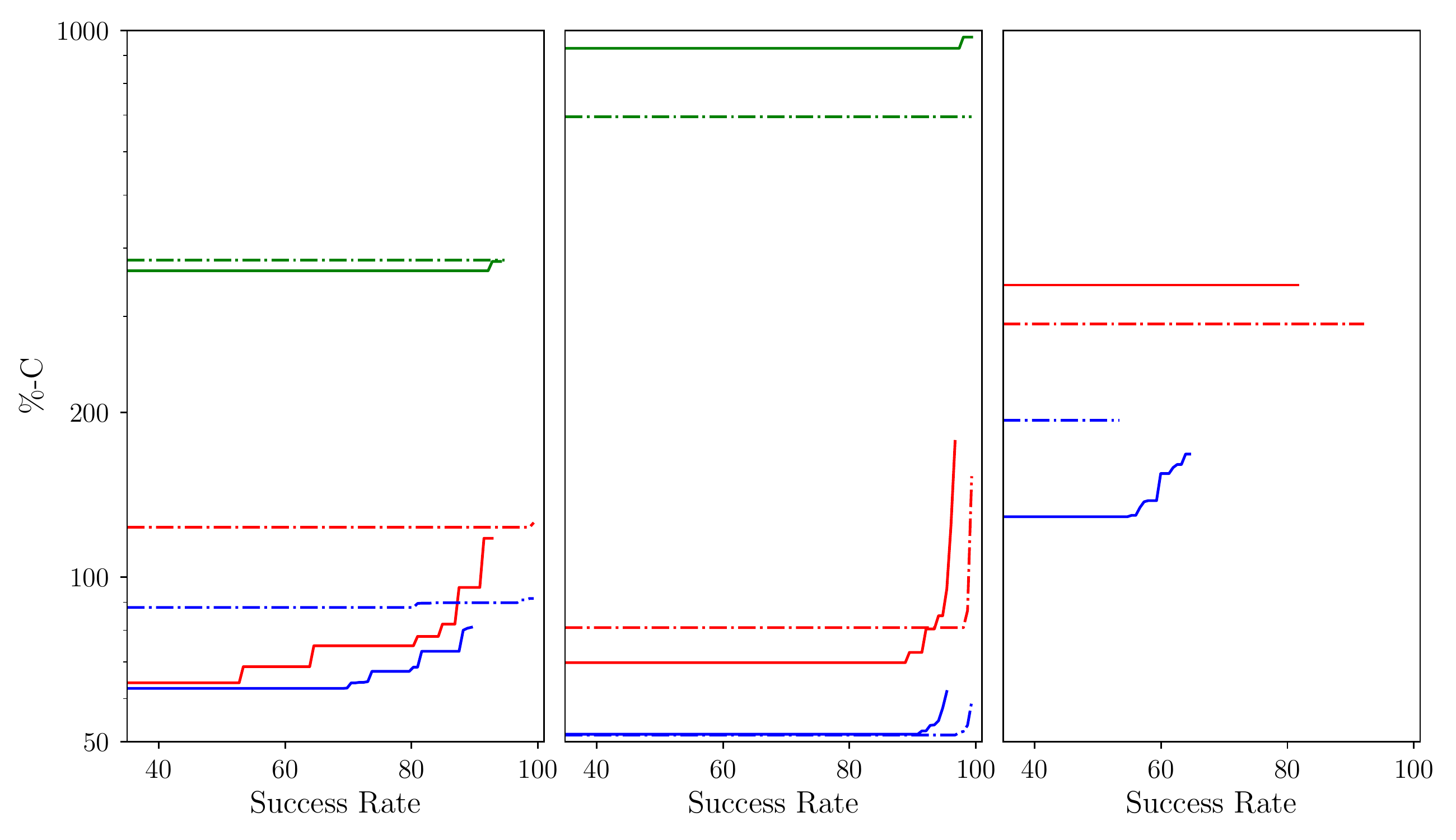}
\end{center}      
\caption{Minimum achievable average percentage difference between the attack radii and the certified guarantee of Cohen \etal (Equation~\ref{eqn:Cohen_Bound}) for a given success rate for our technique (Blue), PGD (Red), and Carlini-Wagner (Green), when tested against MNIST, CIFAR-$10$ and Imagenet. Solid and dashed lines represents $\sigma=\{0.5, 1.0\}$ for parameter space of Table~\ref{tab:parameter_table}. 
}
\label{fig:tradeoff} 
\end{figure*}

\subsection{Attacking Randomised Smoothing}

\paragraph{Attack Size vs Success Rate} %

To establish the performance of our Certification Aware Attack framework against certifications employing randomised smoothing, Figure~\ref{fig:tradeoff} explores the relationship between success rates and the size of the identified adversarial examples. As each technique may be successfully attacking a different subset of samples, the size of the identified adversarial examples is normalised by their associated certified radii in $\%$-C to control for the difficulty of identifying examples. This was achieved through a broad sweep of each model's respective parameter spaces, as detailed within Table~\ref{tab:parameter_table} and Appendix~\ref{sec:parameter_space}. Exemplars of some of these attacks can be seen in Appendix~\ref{sec:examplar}.

These results demonstrate the existence of a quasi-exponential relationship between the attack size and the location of the smallest \emph{potential} adversarial example, which serves as a proxy for detectability. This growth is in part a product of the range of the explored parameter space (see Appendix~\ref{sec:parameter_space}), which extends to attacks so large that they remove all semantic features from the inputs---producing almost guaranteed attack successes, at the cost of clearly detectable perturbations. 

Our approach consistently identifies significantly smaller adversarial examples than both PGD and Carlini-Wagner for all success rates, with a $20$ percentage point difference in the distance to Cohen \etal seen for CIFAR-$10$ at $\sigma=1.0$, and an over $30$ percentage point difference for MNIST and ImageNet. While it is true that our technique produces slightly smaller maximum success rates than PGD in MNIST and CIFAR-$10$ (with a larger, notable gap for ImageNet), we emphasise the significant differences between the observed percentages distances to Cohen, especially as the success rate grows for MNIST and CIFAR-$10$. The the  adversarial examples identified by our technique are consistently closer to the certified radii confirms that our technique is reliably producing smaller, more difficult to detect adversarial examples.

\paragraph{Relative Performance} To enable representative comparisons, for the remainder of this work we will assume that an attacker that can control its position in parameter space will choose hyper-parameters such that it minimises the median percentage difference to Cohen \etal for a success rate above $90 \%$. If such a success rate is not achievable, the attacker will instead maximise their success rate.

To more comprehensively examine the suite of potential attack frameworks, we now expand our suite of comparisons to also include DeepFool \citep{moosavi2016deepfool} and AutoAttack \citep{croce2020reliable}, which were excluded from broader parameter sweeps due to their relative performance. While AutoAttack has the ability to specify a $\ell_2$ norm perturbation magnitude, the associated computational cost makes a broader parameter space exploration infeasible. In contrast, while DeepFool is the fastest of all tested attacks, its failure to successfully identify norm minimising adversarial examples led to its exclusion from a broader parameter exploration.

Across our full set of experiments, Figure~\ref{fig:consolidated} and Table~\ref{tab:main_times} demonstrate that our new attack framework consistently constructs smaller adversarial perturbations than any other tested technique. On a sample-by-sample basis, in the most challenge experiment for our technique---Imagenet at $\sigma = 1.0$---our technique produces the smallest adversarial example for $54 \%$ of the time (denoted by the \emph{Best} column), for samples able to be attacked. This result is particularly striking given the relatively low success rate for our approach in Imagenet, relative to the other experiments, which suggests that the range of parameter space tested over may need further modification for datasets of the size and complexity of Imagenet. In the remainder of the tested experiments, as the success rate of our technique increases, so too does the proportion of attacked samples for which our technique produces the smallest possible adversarial attack, demonstrating the viability of our approach as a framework for constructing minimal norm adversarial examples. 

Our approach produces a median certification that is on average $11 \%$ smaller for MNIST, $12 \%$ for CIFAR-$10$, and $52 \%$ smaller in the case of Imagenet. When controlling for the size of the certified radii $\%$-C demonstrates that our technique produces an on average $24 \%$ reduction in the median attack size relative to the next best attack.%

\paragraph{Contextualising Performance} One feature noted within Section~\ref{sec:refining} was that  all adversarial examples identified by our Certification Aware Attack framework must share the same class prediction as the first identified adversarial example. Intuitively it would appear that such a drawback would induce a disproportionate increase in the median certified radii for the $1000$-class ImageNet, as compared to MNIST or CIFAR-$10$. In practice this is more than counterbalanced by our attacks increased efficiency in exploring the search space. This efficiency in exploring the search space is evident in the computational cost of identifying attacks, with our approach requiring significantly less computational time to identify norm minimising adversarial examples relative to all of the other techniques. The exception to this is DeepFool, however its performance delta relative to the Cohen \etal certified radius emphasises that adding additional iterative steps would likely be a forlorn task.

\begin{table*}
  \caption{MNIST (M), CIFAR-$10$ (C), and ImageNet (I) attack performance across $\sigma$ for Carlini-Wagner (C-W), AutoAttack (Auto) and DeepFool (DeepF). Metrics are the proportion of samples attacked (\emph{Suc.}), smallest attack proportion (\emph{Best}), median attack size ($r_{50}$), time (\emph{Time} [s]), and percentage difference to the Cohen \etal ($\%$-C). All bar the success rate are only calculated over \emph{successful attacks}.$\star$ denotes solutions selected following Appendix~\ref{sec:parameter_space}, bolded values represent the best performing metric (excluding the success rate, as it is a control parameter), and arrows denote if a metric is more favourable with increased or decreased values.
  } 
  \label{tab:main_times}
  \centering
  \footnotesize
  \begin{tabular}{llrrS[table-format=2.2]rS[table-format=2.2]|rrS[table-format=2.2]rS[table-format=2.2]}
    \toprule
    \multicolumn{2}{c}{ } & \multicolumn{5}{c}{$\sigma = 0.5$} & \multicolumn{5}{c}{$\sigma = 1.0$} \\
 & Type & Suc.$\uparrow$ & Best 
 $\uparrow$ & $r_{50} \downarrow$ & \%\text{-C}$\downarrow$ & $\text{Time}\downarrow$ & Suc.$\uparrow$ & Best 
 $\uparrow$ & $r_{50} \downarrow$ & \%\text{-C}$\downarrow$ & $\text{Time}\downarrow$ \\
\cmidrule(r){1-2} \cmidrule(r){3-7} \cmidrule(r){8-12}
M & $\text{Ours}^{\star}$ & $90 \%$ & $\mathbf{73 \%}$  & \textbf{2.02} & $\mathbf{82}$ & 0.34 & $97 \%$ & $\mathbf{97 \%}$  & \textbf{2.23} & $\mathbf{90}$ & 1.22 \\
 & $\text{PGD}^{\star}$ & $91 \%$ & $19 \%$  & 2.17 & $96$ & 2.04 & $99 \%$ & $3 \%$  & 2.62 & $123$ & 2.03 \\
 & $\text{C-W}^{\star}$ & $93 \%$ & $7 \%$  & 5.46 & $364$ & 3.03 & $95 \%$ & $0 \%$  & 5.36 & $380$ & 3.02 \\
 & Auto & $92 \%$ & $1 \%$  & 5.44 & $393$ & 27.32 & $97 \%$ & $0 \%$  & 5.65 & $386$ & 26.50 \\
 & DeepF & $9 \%$ & $0 \%$  & 14.43 & $2417$ & \textbf{0.07} & $51 \%$ & $0 \%$  & 17.10 & $2143$ & \textbf{0.07} \\
\cmidrule(r){1-2} \cmidrule(r){3-7} \cmidrule(r){8-12}
C & $\text{Ours}^{\star}$ & $91 \%$ & $\mathbf{87 \%}$  & \textbf{0.83} & $\mathbf{56}$ & 0.53 & $96 \%$ & $\mathbf{92 \%}$  & \textbf{1.26} & $\mathbf{56}$ & 0.86 \\
 & $\text{PGD}^{\star}$ & $92 \%$ & $4 \%$  & 0.92 & $72$ & 2.17 & $99 \%$ & $3 \%$  & 1.46 & $77$ & 2.15 \\
 & $\text{C-W}^{\star}$& $98 \%$ & $5 \%$  & 3.13 & $432$ & 3.18 & $99 \%$ & $1 \%$  & 3.65 & $352$ & 3.14 \\
 & Auto & $94 \%$ & $3 \%$  & 4.00 & $493$ & 28.37 & $91 \%$ & $2 \%$  & 5.61 & $492$ & 28.40 \\
 & DeepF & $88 \%$ & $2 \%$  & 2.44 & $504$ & \textbf{0.08} & $98 \%$ & $3 \%$  & 3.42 & $462$ & \textbf{0.08} \\
\cmidrule(r){1-2} \cmidrule(r){3-7} \cmidrule(r){8-12}
I & $\text{Ours}^{\star}$ & $63 \%$ & $\mathbf{84 \%}$  & \textbf{1.14} & $\mathbf{127}$ & 4.49 & $52 \%$ & $\mathbf{63 \%}$  & \textbf{1.43} & $\mathbf{157}$ & 5.21 \\
 & $\text{PGD}^{\star}$ & $82 \%$ & $13 \%$  & 2.05 & $211$ & 51.46 & $91 \%$ & $35 \%$  & 3.42 & $188$ & 50.49 \\
 & $\text{C-W}^{\star}$& $53 \%$ & $0 \%$  & 33.22 & $4747$ & 26.71 & $56 \%$ & $0 \%$  & 32.42 & $3451$ & 27.34 \\
 & DeepF & $56 \%$ & $3 \%$  & 2.89 & $654$ & \textbf{2.88} & $71 \%$ & $3 \%$  & 4.98 & $647$ & \textbf{2.93} \\
    \bottomrule
  \end{tabular}
\end{table*}

\subsection{Tightness of Certified Guarantees} 

While attacks against certifications are valuable in their own right, they also serve as an upper-bound on the location of the nearest adversarial example, counterbalancing the lower-bound provided by certification mechanisms. In doing so, they provide a proxy for both utility and the potential scope for future improvements in certification schemes. %

\begin{figure*}
\begin{center}
    \includegraphics[width=0.85\textwidth]  {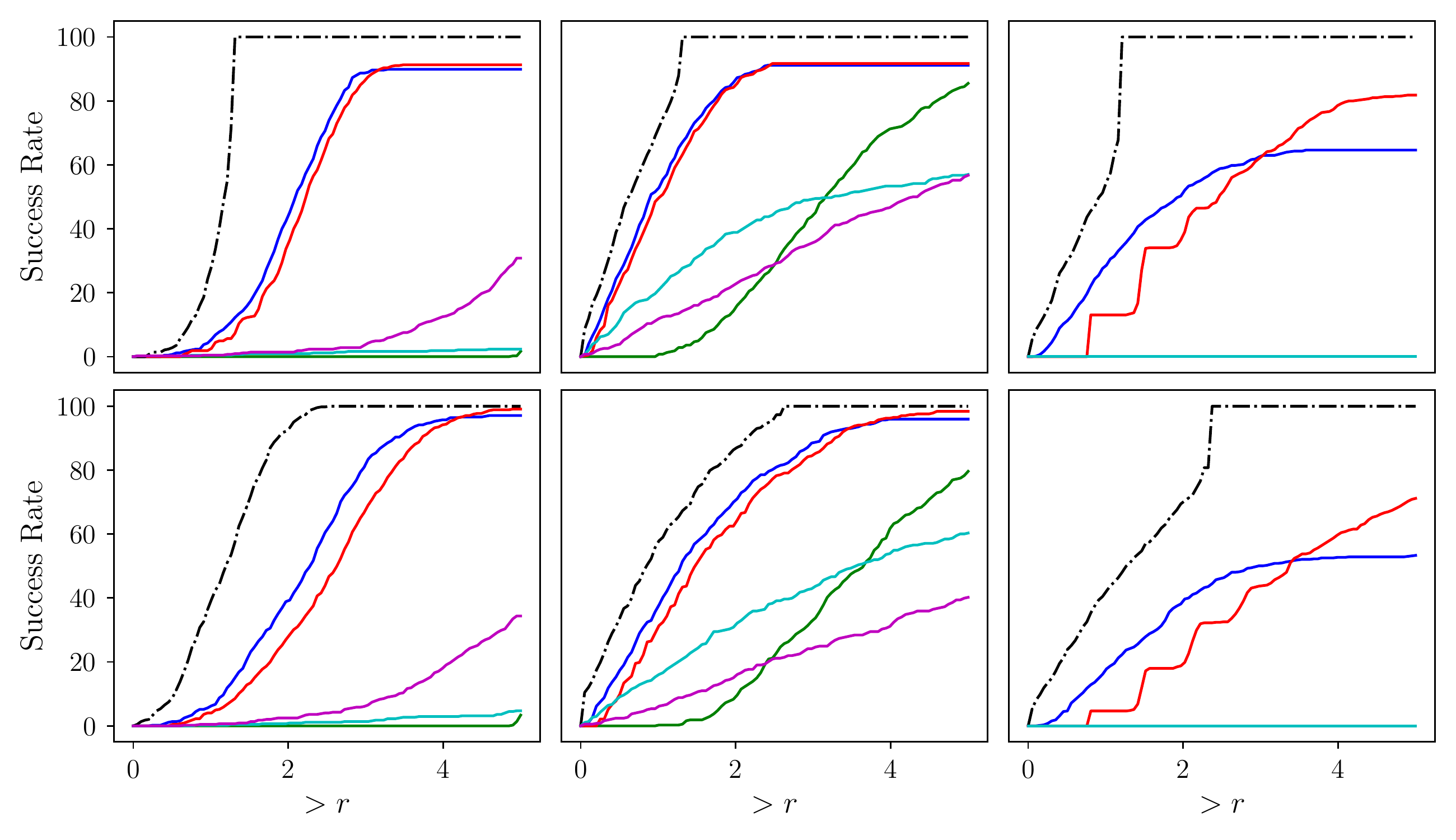}
\end{center}
\caption{Best achieved Attack Proportion for our new Certification Aware Attack (blue), PGD (red), DeepFool (cyan), Carlini-Wagner (green), and AutoAttack (magenta); where the rows correspond to $\sigma = \{0.5, 1.0\}$ and the columns correspond to MNIST, CIFAR-$10$ and Imagenet. Black dotted line represents the best case performance as per Equation~\ref{eqn:Cohen_Bound}.}
\label{fig:consolidated} 
\end{figure*}

When considering these bounds in the contexts of the tested datasets, Figure~\ref{fig:consolidated} suggests that MNIST---which is often perceived as being the simplest of datasets---demonstrates a more significant delta between the certified radii and the attack performance. This may be a consequence of the simpler semantic properties of the dataset being more difficult to attack, relative to Cohen \etal style certifications. Due to its role as a multiplicative constant in Equation~\ref{eqn:Cohen_Bound}, increasing $\sigma$ inherently increases the size of the certifications, an effect that is partially offset by a decrease in the observed class expectations. However, from the perspective of an attacker increasing $\sigma$ should also increase the smoothness of the gradients, which theoretically should make the model \emph{easier to attack}. In practice, Figure~\ref{fig:consolidated} and Table~\ref{tab:main_times} demonstrate that increasing $\sigma$ leads to a small increase in the size of identified attacks, relative to certified guarantees. While this may at first appear contradictory, it suggests that the ease in identifying adversarial attacks for larger $\sigma$ is offset by decreases in the tightness of the certified bound.

\subsection{Performance against other certification mechanisms}\label{sec:attacking_ibp}

\begin{figure}[!htbp]
    \centering
    \includegraphics[width=.8\linewidth]{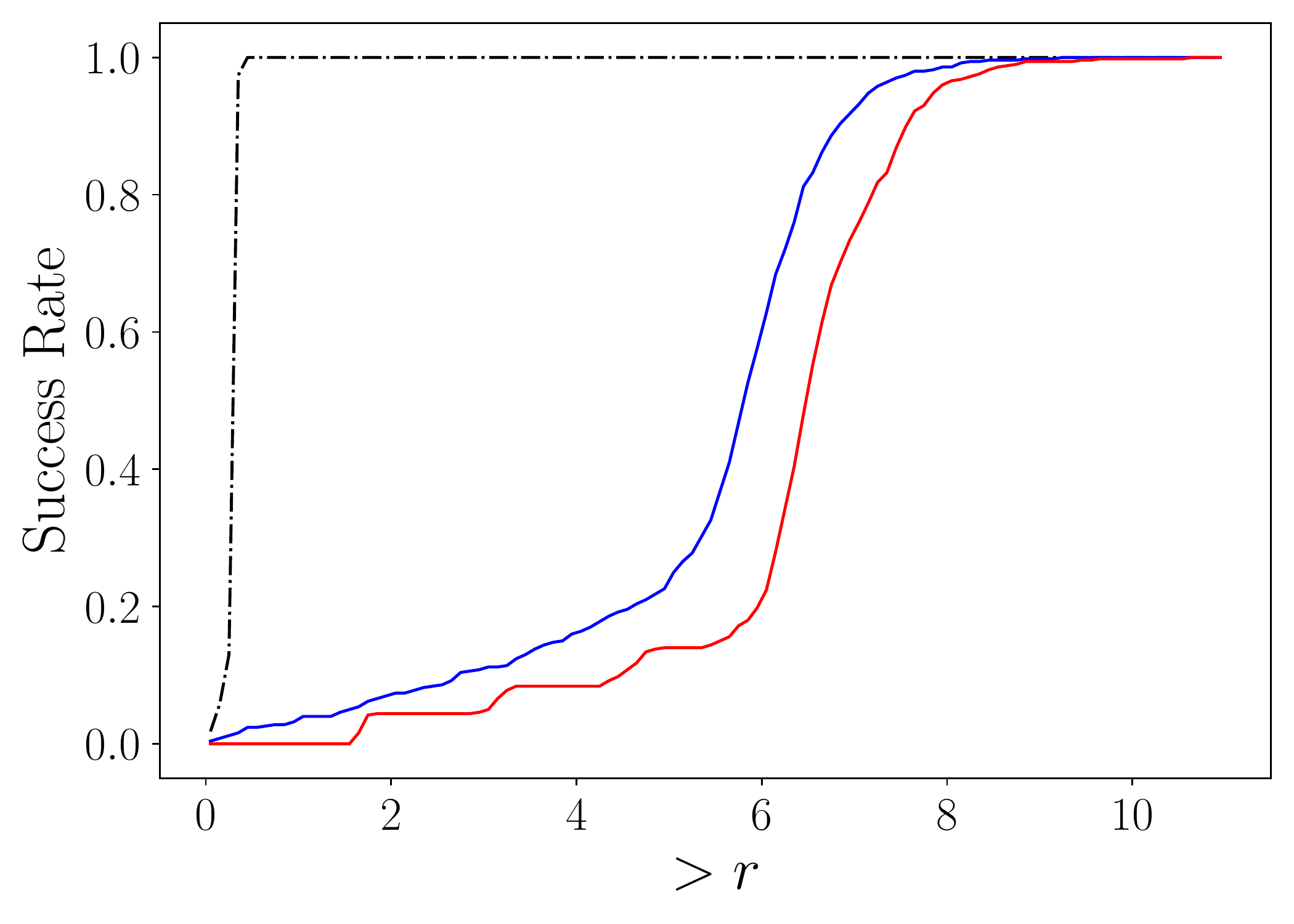} 
  \caption{Success rates for our attack (blue) and PGD (red) for an IBP certified MNIST model.}
  \label{fig:IBP}
\end{figure}

To demonstrate the generality of our identified threat model, Figure~\ref{fig:IBP} demonstrates the relative performance of our technique and PGD when tested for a model certified using IBP. While we have not fully explored the parameter spaces of both attacks, nor the broader suite of attacks in the context of this framework, these results reinforce the \emph{information advantage} an attacker has when attempting to compromise models employing randomised smoothing \emph{if they incorporate the certification into their attack}, irrespective of the certification mechanism. That this is true confirms that all certification mechanisms should assess their risk to adversarial attack in light of our Certification Aware Attacks.

\section{Discussion: Impact and Mitigation}

We emphasise that our CAA does not compromise the certification associated with individual samples. Rather, we establish that it is possible to leverage certifications to construct samples that are semantically identical to clean samples (see Figure~\ref{fig:illustrative_examples}), but that are able to trick a certified classifier into changing its class prediction. That this is possible demonstrates that the security provided by certification mechanisms is illusory if the attacker knows (or can reconstruct) the robustness certificate. This observation runs contrary to likely user expectations, where a certification or the class expectations is likely to be be seen as metric for demonstrating model confidence to be released alongside the predicted class.

While uncovering new attacks has the potential to compromise deployed systems, there is a prima facie argument that any security provided by ignoring new attack vectors is illusory. Such a perspective has uncovered new attack vectors including data poisoning, backdoor attacks, model stealing, and transfer attacks, all of which can now be protected against. Our work similarly reveals the paradoxical observation that the very mechanisms that we rely upon to protect models introduce new attack surfaces, enhancing the ability for attackers to construct norm minimising attacks. Moreover, our CAA also provides a framework to better explore the tightness of constructed certification bounds in real world systems. 

It would appear that restricting the publication of the certification could nullify the potential of certification aware attacks. However, the certifications can be trivially reconstructed from the class expectations and $\sigma$. Thus securing systems against these attacks requires that only the  class predictions are released, without any associated expectations. While this would not prevent a suitably motivated attacker developing a surrogate model, it would significantly increase the cost and difficulty associated with a successful attack, relative to attacking a model that releases its certifications. Figure~\ref{fig:app:sigma} and Appendix~\ref{app:sigma_accuracy} demonstrate that our attack still outperforms other frameworks even when $\sigma$ is estimated, however we leave exploring the effectiveness of this mitigation to future work. %

While this work has specifically considered $\ell_2$ norm bounded evasion attacks against randomised smoothing and IBP based mechanisms, it is important to note that other threat models threat models~\citep{liu2023enhancing}, and certification frameworks exist, like Lipschitz certification~\cite{tsuzuku2018lipschitz, leino2021globally}. However, these approaches still involve constructing a certificate, and if these certificates are published then this work has clearly demonstrated that this release can be exploited by a motivated attacker. As such, we believe that any future certification---for evasion, backdoor, or other attacks---must seriously consider the risk associated with publishing certifications.

\section{Related Work}

This work presents both a framework for attacking certifiably robust models, and a demonstration of how such certification can be exploited to improve attack efficacy. While we formalise the concept of attacking certifications, prior works have considered the impact of corrupting the inputs of both undefended and certified models. One common framework involves corrupting input samples with additive noise or adversarial examples, in order to improve robustness~\citep{bishop1995training, salman2019provably, cohen2019certified}. Of these, \citet{salman2019provably} is closest to our work, although their attacks only considered a small number of draws from randomised smoothing (rather than the full expectations), and employed a $\softmax$ in place of the $\argmax$ operator. All three of these approaches are un-targeted, un-directed, training time modifications attempting to improve generalisation by increasing training loss. In contrast, our focus was placed upon both constructing a definition of test time adversarial attacks against certified models, and then exploiting the nature of certifications themselves to improve the performance of adversarial attacks against certified models.

\section{Conclusion}

Within this work we have demonstrated the counter-intuitive concept that certifications can be exploited to attack the very models they were designed to guard. Through our novel Certification Aware Attack framework, we exploit this observation to significantly decrease the size of the identified adversarial perturbations relative to state-of-the-art test-time attacks, leading to an up to $55 \%$ decrease in the size of adversarial perturbations relative to the next best performing technique. Being able to reliably, and repeatedly generate such norm-minimising adversarial examples would better allow for developers to analyse the performance of certification mechanisms. However, this same benefit would also allow an attacker to reliably influence more samples before potentially being detected. These results underscore that significant consideration must be placed upon the safety of releasing robustness certificates. %

\newpage

\section*{Impact Statement}

This work explores a heretofore undiscovered security risk associated with publishing the certifications associated with certifiable robust mechanisms. While such attacks can induce negative societal consequences, there is a clear consensus within the computer security community that responsible disclosure of attacks is a crucial part of improving systemic security. Training-time, test-time and backdoor attacks are always published in the public domain, and the risk associated with the publication of our new attack should be viewed within this context. The new attack vector contained within this work allows for a fundamental reevaluation of the security associated with systems designed to demonstrate the resistance of machine learning models to adversarial manipulation, and provides a new mechanism that can be used to better study the tightness of the bounds provided by robustness mechanisms. 

\bibliography{ICML2022_bib}
\bibliographystyle{iclr2023_conference}

\appendix

\section{Background: Adversarial Examples}\label{app:adversarial_examples}

The existence of highly confident but incorrect adversarial examples in neural networks has been documented  extensively~\citep{szegedy2013intriguing, goodfellow2014explaining}. We provide an overview of the topic in this appendix for completeness. Formally, adversarial examples are perturbations $\boldsymbol{\gamma}\in\mathcal{S}$ to the input $\mathbf{x}\in\mathcal{S}$ of a learned model $\mathbf{f}(\cdot)$, for which 
    $F(\mathbf{x} + \boldsymbol{\gamma}) \neq
    F(\mathbf{x})$.

The $p$-norm of this perturbation can be considered a reliable proxy for both the \emph{detectability} of adversarial examples~\citep{gilmer2018motivating} and the \emph{cost} to the attacker~\citep{huang2011adversarial}.

The process for identifying such attacks commonly involves gradient descent over the input space. A prominent example is the Iterative Fast Gradient Sign Method \citep{madry2017towards, dong2018boosting}, which we will henceforth refer to as PGD. This technique attempts to converge upon an adversarial example by way of the iterative scheme
\begin{equation}\label{eqn:PGD_its}\mathbf{x}_{k+1} = P_{\mathcal{S}} \left( \mathbf{x}_{k} - \epsilon \left(\frac{\nabla_{\mathbf{x}} J(\mathbf{x}, y)}{\norm{\nabla_{\mathbf{x}} J(\mathbf{x}, y)}_{2}}\right) \right)\enspace.\end{equation}
This process exploits gradients of the loss $J(\mathbf{x}, y)$ relative to a target label $y$ to form each attack iteration, with the step size $\epsilon$ and a projection operator $P$  ensuring that $\mathbf{x}_{k+1}$ is restricted to $\mathcal{S}$. %

\citet{carlini2017towards}---henceforth known as C-W---demonstrated the construction of adversarial perturbations by employing gradient descent to solve
\begin{align}\label{eqn:CW}
    &\argmin_{\mathbf{x}'}\left\{ \norm{\mathbf{x}' - \mathbf{x}}_{2}^{2} + 
      c \cdot g
    \right\}\\
    & g =  \max \left\{ \max \{ f_{\theta}(\mathbf{x}')_{j} : j \neq i\} - f_{\theta}(\mathbf{x}')_{i}, -\kappa \right\} 
    \vphantom{ \norm{\mathbf{x}' - \mathbf{x}}_{2}^{2} } \enspace. \nonumber %
\end{align}

We note that while one-shot variants of these attacks have historically been used as a baseline for the performance of iterative attacks to be assessed against, we believe that by their nature such attacks always poorly represent the success-rate and attack-size trade off. Instead, we have performed our comparisons against the certified guarantee of Cohen \etal at the sample point, which provides an absolute lower bound on the size of possible adversarial attacks. We feel that this form of comparison more appropriately captures how these techniques perform, rather than attempting to compare one-shot with iterative attacks, which fundamentally incorporate different access level threat models.

\section{Algorithms}\label{app:algorithms}

Within Algorithm~\ref{alg:CAA}, lines $6$--$11$ cover the processes outlined within Section~\ref{sec:CAA} and \ref{sec:specifying}, with lines $13$--$17$ covering the materials of Section~\ref{sec:refining}. 

One important piece of detail relates to the case where $\widecheck{E}_0 < \widehat{E}_1$, which is equivalent to $r=0$. Under both of these circumstances, the model is unable to construct a confident prediction, so the algorithm induces minimal size-steps either away from the origin---if an adversarial example has not yet been identified---or towards the most recent point, if that point was an adversarial example. 

\begin{algorithm}[tb]
   \caption{Certification Aware Attack Algorithm.}%
   \label{alg:CAA}
\begin{algorithmic}[1]
   \State {\bfseries Input:} data $\mathbf{x}$, level of additive noise $\sigma$, samples $N$, iterations $M$, true-label $i$, minimum and maximum step size $\left(\epsilon_{\text{min}}, \epsilon_{\text{max}} \right)$, scaling factor $\delta \in [0,1]$
   \State $\mathbf{x}', \mathbf{x}'_{s} $ Successful $ = \mathbf{x}, \mathbf{x}, $ False
   \For{$1$ {\bfseries to} $M$}
   \State $\mathbf{y}, \widecheck{E}_0, \widehat{E}_1, r = \text{Model}(\mathbf{x}'; \sigma, N)$ \algorithmiccomment{Detailed in Algorithm~\ref{alg:model_predict}}
   \If{$\argmax_{i \in \mathcal{K}}{y} = i$} \algorithmiccomment{Adversarial Example not yet identified.}
   \If{$\widecheck{E}_0 > \widehat{E}_1$}
   \State $\epsilon = $ Equation~\ref{eqn:epsilon_basic} $(\mathbf{x}', \delta,  \epsilon_{\text{min}},  \epsilon_{\text{max}})$
   \Else
   \State $\epsilon = \epsilon_{\text{min}}$
   \EndIf
   \State $\mathbf{x}' = $ Equation~\ref{eqn:CAA_iter} $(\mathbf{x}', \epsilon)$ 
   \Else
   \If{$r = 0$}  \algorithmiccomment{Attempting to improve confidence of adversarial examples} 
        \State $\mathbf{x'} = P_{\mathcal{S}}\left(\mathbf{x'} + \epsilon_{\text{min}} \frac{\nabla_{\mathbf{x}'} \left( \widecheck{E}_0 - \widehat{E}_1 \right)}{\|\nabla_{\mathbf{x}'} \left( \widecheck{E}_0 - \widehat{E}_1 \right)\|_2}\right)$ 
   \Else \algorithmiccomment{Examples are refined while staying inside the certified radii}
   \State $\mathbf{x}'_s, $ Successful = $\mathbf{x}', $ True
   \State $\mathbf{x}' = $ Equation~\ref{eqn:correction}$(\mathbf{x}', \delta,  \epsilon_{\text{max}})$ 
\EndIf
\EndIf
   \EndFor  
   \State \textbf{return} $\mathbf{x}'_s$, Successful
\end{algorithmic}
\end{algorithm}

In order to calculate the class expectations and associated certifications for a given input $\mathbf{x}'$, Algorithm~\ref{alg:model_predict} performs the Monte-Carlo sampling and then corrects for sampling uncertainties. We note here that for the purposes of constructing derivatives, the lower and upper-bounding processes are treated as if they were perturbations to the expectations, and as such they are not considered as a part of the differentiation process. While this has the potential to slightly perturb the derivatives, our experiments have demonstrated that any $\delta > 0$ is sufficient to more than compensate

\begin{algorithm*}[tb]
   \caption{Class prediction and certification for the Certification Aware Attack algorithm of Algorithm~\ref{alg:CAA}.}%
   \label{alg:model_predict}
\begin{algorithmic}[1]
   \State {\bfseries Input:} Perturbed data $\mathbf{x}'$, samples $N$, level of added noise $\sigma$
   \State $\mathbf{y} = \mathbf{0}$
   \For{i = 1:N}
   \State $\mathbf{y} = \mathbf{y} + GS \left( f_{\theta}\left(\mathbf{x}' + \mathcal{N}(0, \boldsymbol{\sigma}^2) \right)\right)$
   \EndFor
   \State $\mathbf{y} = \frac{1}{N} \mathbf{y}$
   \State $(z_0, z_1) = \topk (\mathbf{y}, k=2)$ \algorithmiccomment{$\topk$ is used as it is differentiable, $z_0 > z_1$}
  \State $\left(\widecheck{E}_0, \widehat{E}_1\right) = \left(\lowerbound (\mathbf{y}, z_0), \upperbound (\mathbf{y}, z_1)\right)$ \algorithmiccomment{Calculated via  \citet{goodman1965simultaneous}}
  \State $R = \frac{\sigma}{2} \left(\Phi^{-1}(\widecheck{E}_0) - \Phi^{-1}(\widehat{E}_1) \right)$ %
  \State \textbf{return} $\mathbf{y}, \widecheck{E}_0, \widehat{E}_1, R$
\end{algorithmic}
\end{algorithm*}

\section{Parameter Space}\label{sec:parameter_space}

As was discussed in Section~\ref{sec:results}, understand the relative performance of techniques requires a consideration of how an attacks parameter space influences its performance metrics. In aide of this, for our three most highly performant attack frameworks, for each dataset we performed a parameter sweep over the parameters outlined within Figure~\ref{tab:parameter_table}. From this, for each attack we selected a representative position in parameter space that either exhibited the minimal $\%\text{-C}$ for a success rate over $90 \%$, or, if such a success rate was not achievable, the maximum achievable success rate. In doing so, we attempted to construct fair comparisons that accurately reflected the performance of the techniques. 

We note that within this table the parameter $\epsilon_{\text{max}}$ (for our technique) and $\epsilon$ for PGD extend to a level close to that for which the semantic features of the inputs would entirely be destroyed by the attacker. While this choice was made in the interests of validating the performance of our attacks, we emphasise that none of these results ended up factoring into our key reported results. 

One complicating factor of such parameter sweeps is the computational cost associated with the exploration, especially in the case for Imagenet---as can be seen in Table~\ref{tab:main_times}. As such while we endeavoured to select our representative attacks based upon $500$ randomly selected samples, it was only possible to consider $50$ samples for PGD and Carlini-Wagner for Imagenet due to the computational time associated with these parameter sweeps. 

\begin{table*}[]
    \caption{Parameter space employed for our Certification Aware Attack, PGD (see Equation~\ref{eqn:PGD_its} for details), and Carlini-Wagner (see Equation~\ref{eqn:CW}). }
    \label{tab:parameter_table}
    \centering
    \begin{tabular}{l|rcl}
    \toprule
       Ours & $\epsilon_{\text{min}} \times 255$ & $=$ & $\{1, 5, 10\}$\\
         & $\epsilon_{\text{max}} \times 255$ & $=$ & $\{20, 40, 100, 255\}$\\
         & $\delta$ & $=$ & $\{0.01, 0.025, 0.05, 0.075, 0.1\}$ \\
      PGD & $\epsilon \times 255$ & $=$ & $\{1, 4, 8, 10, 20, 30, 40, 50, 100, 200\}$\\
      C-W & $c $ & $=$ & $\{ 10^{-5}, 10^{-4}, 10^{-2}, 10^{-1}, 1, 2, 3\}$\\
      \bottomrule
    \end{tabular}
\end{table*}

\begin{figure*}
    \centering
    \includegraphics[width=.75\linewidth]{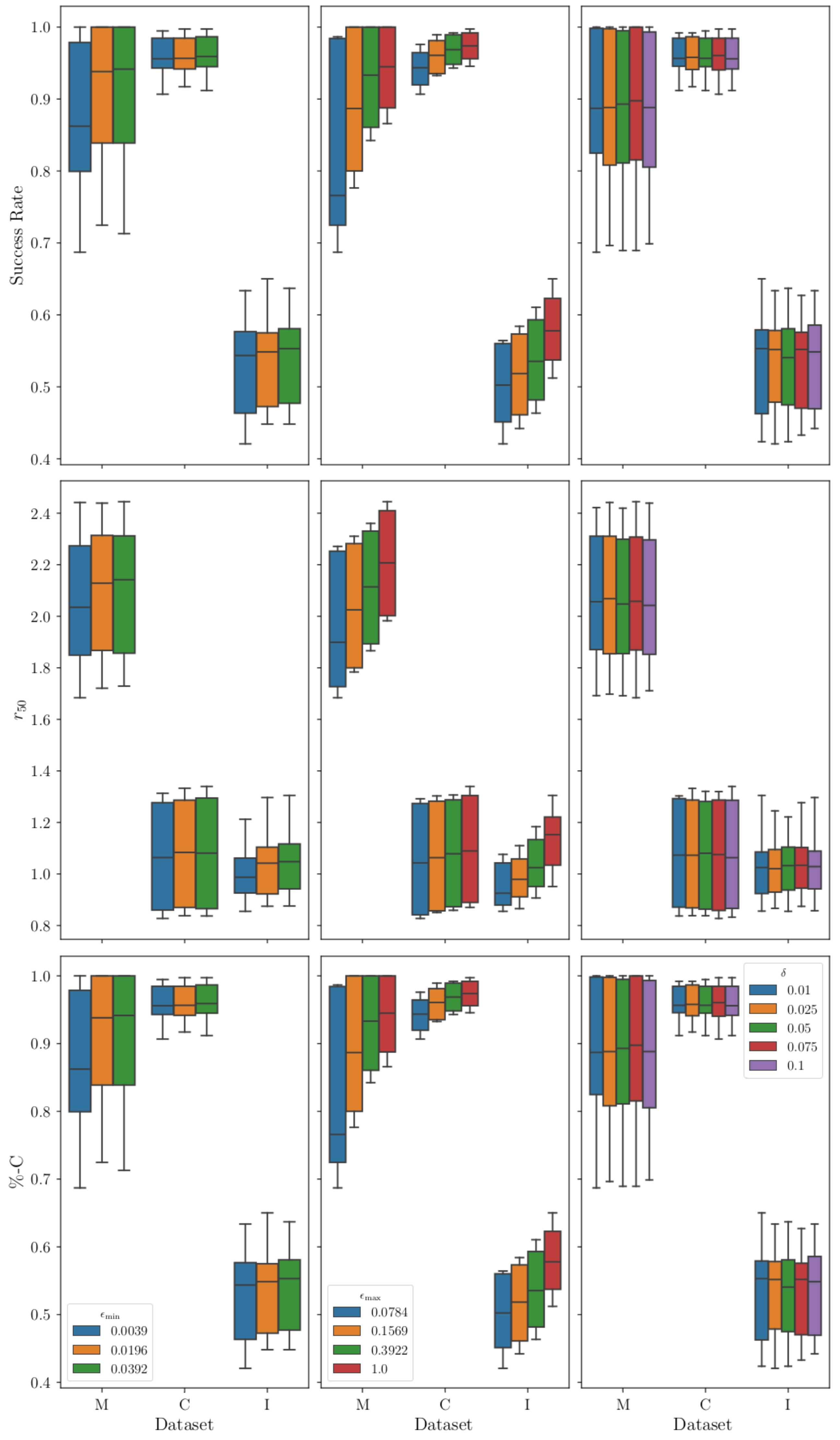}
  \caption{Response of key metrics for our Certification Aware Attack to changes in $\epsilon_{\text{min}}$, $\epsilon_{\text{max}}$ and $\delta$.}\label{fig:our_parameter_space}   
\end{figure*}

To explore the influence of the step-size control parameters of Equation~\ref{eqn:step_size_limits}, Figure~\ref{fig:our_parameter_space} considers the influence of a range of these parameters upon key attack metrics, based upon the parameter space explored over Appendix~\ref{sec:parameter_space}. Based upon this , it is clear that the primary driver of the success-rate and certification size trade off (as explored in Figure~\ref{fig:tradeoff}) is the parameter $\epsilon_{\text{max}}$, that controls the largest possible step size that the Certification Aware Attack framework is allowed to make. Thus further exploring the parameter space in this direction would likely be a critical factor in increasing the success rate observed for Imagenet. 

\section{Metrics}\label{sec:metrics}

To help explore the relative performance of the tested techniques we consider a series of metrics which, in aggregate, reflect the overall performance of the technique. To explain these metrics in additional detail, the \emph{Success Rate} represents the proportion of correctly predicted samples for which a technique is able to construct a successful attack, and can be calculated as
\begin{equation}
\text{Suc.}_i = \frac{1}{N} \sum_{j=1}^{N} (r_{i, j} > 0)\enspace.
\end{equation}
Here the subscript $i$ denotes a particular attack drawn from the set of attacks $\mathcal{I}$, and $r_{i,j} = \| \mathbf{x}_j' - \mathbf{x}_j\|$ is the attack radii, which for notational simplicity is set to $0$ in the case of a failed attack. The set of samples (of size $N$) has been filtered to ensure that each is correctly predicted by the model in the absence of an adversarial attack. 

The \emph{Best} is then the proportion of samples that a particular technique produces an attack radii smaller than any other correctly identified adversarial attack is calculable as
\begin{equation}
    \text{Best}_i =  \frac{\sum_{j=1} r_{i, j} \leq r_{i', j} \qquad \forall \text{ } (i' \neq i) \in \mathcal{I}}{\sum_{j=1} r_{i, j} > 0 \qquad \forall \text{ } i \in \mathcal{I}}
\end{equation}
Increases to both of these metrics are advantageous, although as was noted in Appendix~\ref{sec:parameter_space} each result within Table~\ref{tab:main_times} must be contextualised against the decision to attempt to control the success rate to approximately $90 \%$, if such an success rate was achievable for the technique in light of the tested parameter space. 

The measure $\%$-C represents the median percentage difference between the attack radii and the certified guarantee of Cohen \etal, which takes the form
\begin{equation}
    \%\text{-C} = \text{med}_{r_{i,j} > 0}\left( \frac{r_{i, j} - C(\mathbf{x}_j)}{C(\mathbf{x}_j)} \right)\enspace.
\end{equation}
Here $\text{med}(\cdot)$ is the median over the set of successfully attacked samples, and $C(\mathbf{x}_j)$ is the certified radii for an $\ell_2$ norm, as calculable by Equation~\ref{eqn:Cohen_Bound}. Beyond this, $r_{50}$ is the median certified radii of the samples able to be successfully attacked by a given technique, and Time represents the median attack time (in seconds) across all tested samples. All three of these latter metrics demonstrate favourable performance with decreasing values. 

This broad set of metrics was deliberately chosen to reflect different aspects of performance. However, we call particular attention to $\%$-C, as it is a measure of the size of the adversarial examples \emph{relative to the location of the minimal possible adversarial example}---with the certification of Cohen \etal essentially providing what is in essence characteristic scale that can be used for normalisation. We emphasise that such a measure of relative importance is important to further illuminate performance in light of the fact that the other metrics may not all strictly consider the same samples, as they are often constructed over the set of samples an attack method is successfully able to manipulate.

\section{Samplewise Performance}

\begin{figure*}[!ht]
\centering
   \subfloat[Percentage differences relative to $E_0$ \label{fig:comp_scatter}]{%
      \includegraphics[width=0.36\textwidth]{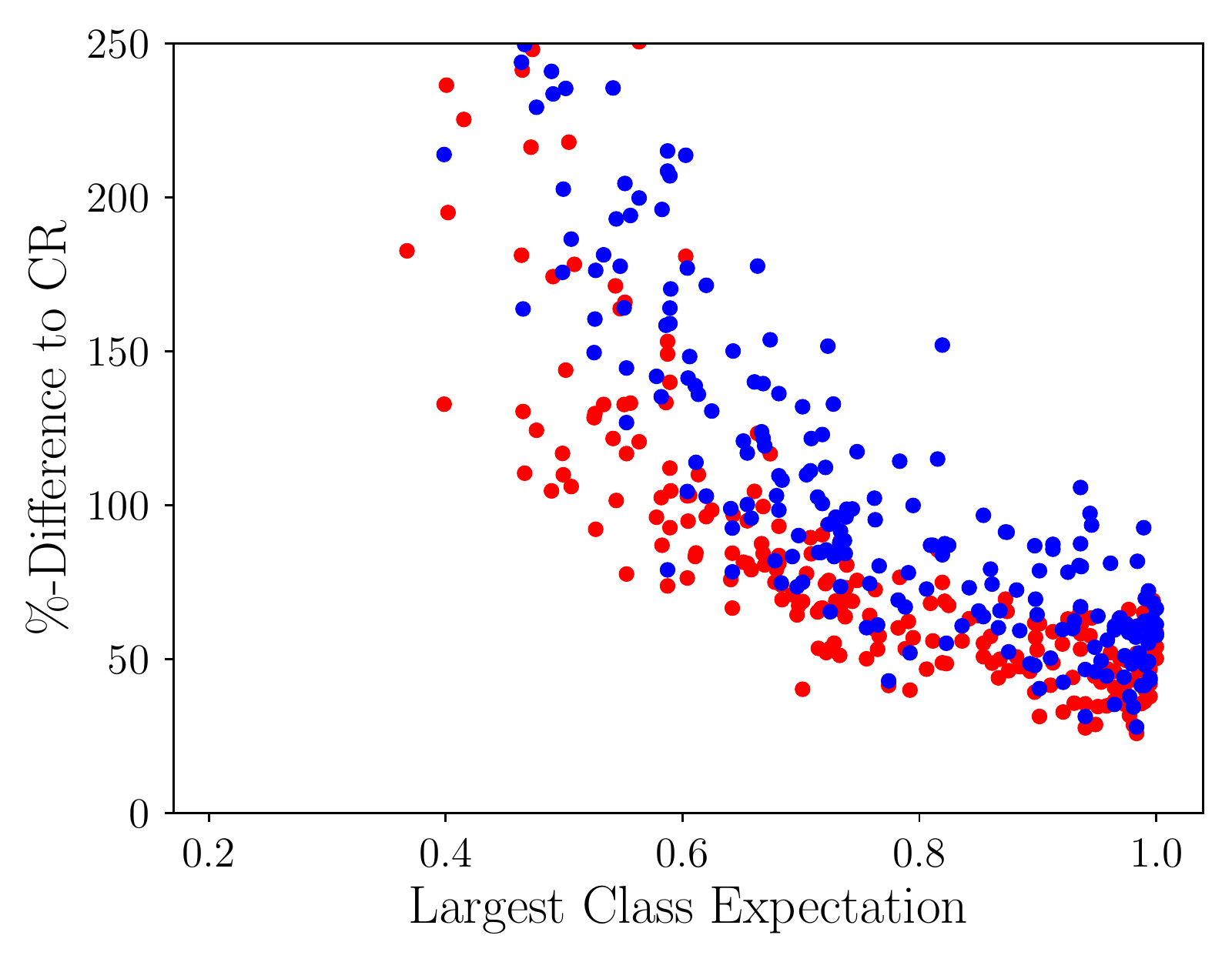}}
   \hspace{1em}
   \subfloat[Performance for estimated $\sigma$ \label{fig:app:sigma}]{%
      \includegraphics[width=0.36\textwidth]{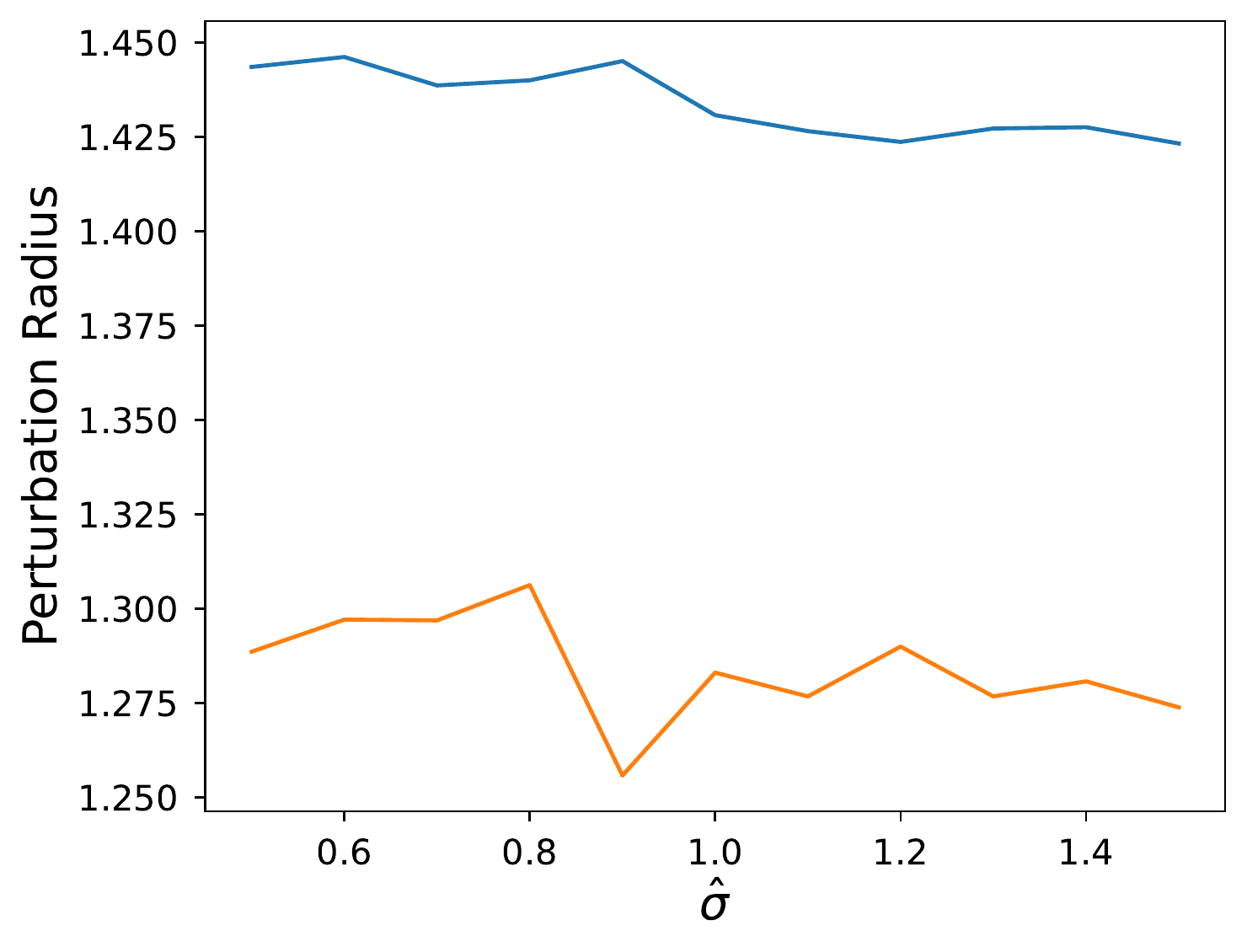}}
\caption{(a) captures the percentage difference between constructed adversarial perturbations and the certified radii of Equation~\ref{eqn:Cohen_Bound} for CIFAR-$10$ at $\sigma=0.5$, with Our technique in red and PGD in blue. (b) demonstrates that the blue mean and orange median performance of our technique are consistent even when $\sigma = 1.0$ is approximated by an estimated $\hat{\sigma}$.}
\end{figure*}

To further illuminate the nature of the performance of our attack, Figure~\ref{fig:comp_scatter} considers the sample-wise performance of both PGD and our Certification Aware Attack. Within this data there is a clear self-similar trend, in which the percentage difference to Equation~\ref{eqn:Cohen_Bound} increases as the largest class expectation decreases. This difference could indicate the potential for improving the certification of samples within this region. There also appears to be a correlation between the outperformance of our approach and the semantic complexity of the prediction task, which suggests that tightening these guarantees could be increasingly relevant for complex datasets of academic and industry interest.

\section{Accuracy of $\sigma$}\label{app:sigma_accuracy}

The white-box threat model assumes that the attacker has access to the full model and its parameters, including the level of additive noise $\sigma$. However, if the attacker only had access to the model and output class expectations, but was somehow prevented from directly accessing $\sigma$ and $r$, it turns out that the Certification Aware Attack can still be applied subject to a sufficiently accurate guess of $\sigma$. As is shown by Figure~\ref{fig:app:sigma}, even over-estimating $\sigma$ by $50 \%$ can decrease the radius of the identified adversarial perturbation under certain experimental conditions. That this is possible is a product of the terms $\delta_1$ and $\delta_2$ in Algorithm~\ref{alg:CAA}, as both of these parameters set the idealised step size to try and either change or preserve the predicted class. While this does suggest that there is potentially additional scope for optimising $\delta_1$ and $\delta_2$, it also demonstrates the possibility of estimating $\sigma$ as part of a surrogate model, in order to attack within a limited threat mode.

\section{Training with MACER}\label{app:macer}

Recent work has considered how certifications might be improved by augmenting the training objective to maximising the expectation gap between classes~\citep{salman2019provably}. A popular approach for this is MACER~\citep{zhai2020macer}, in which the training loss is augmented to incorporate what the authors dub the $\epsilon$-robustness loss, which reflects proportion of training samples with robustness above a threshold level. In principle such a training-time modification can increase the average certified radius by $10$--$20\%$, however doing so does increase the overall training cost by more than an order of magnitude.

\begin{figure*}
    \centering
    \includegraphics[width=.7\linewidth]{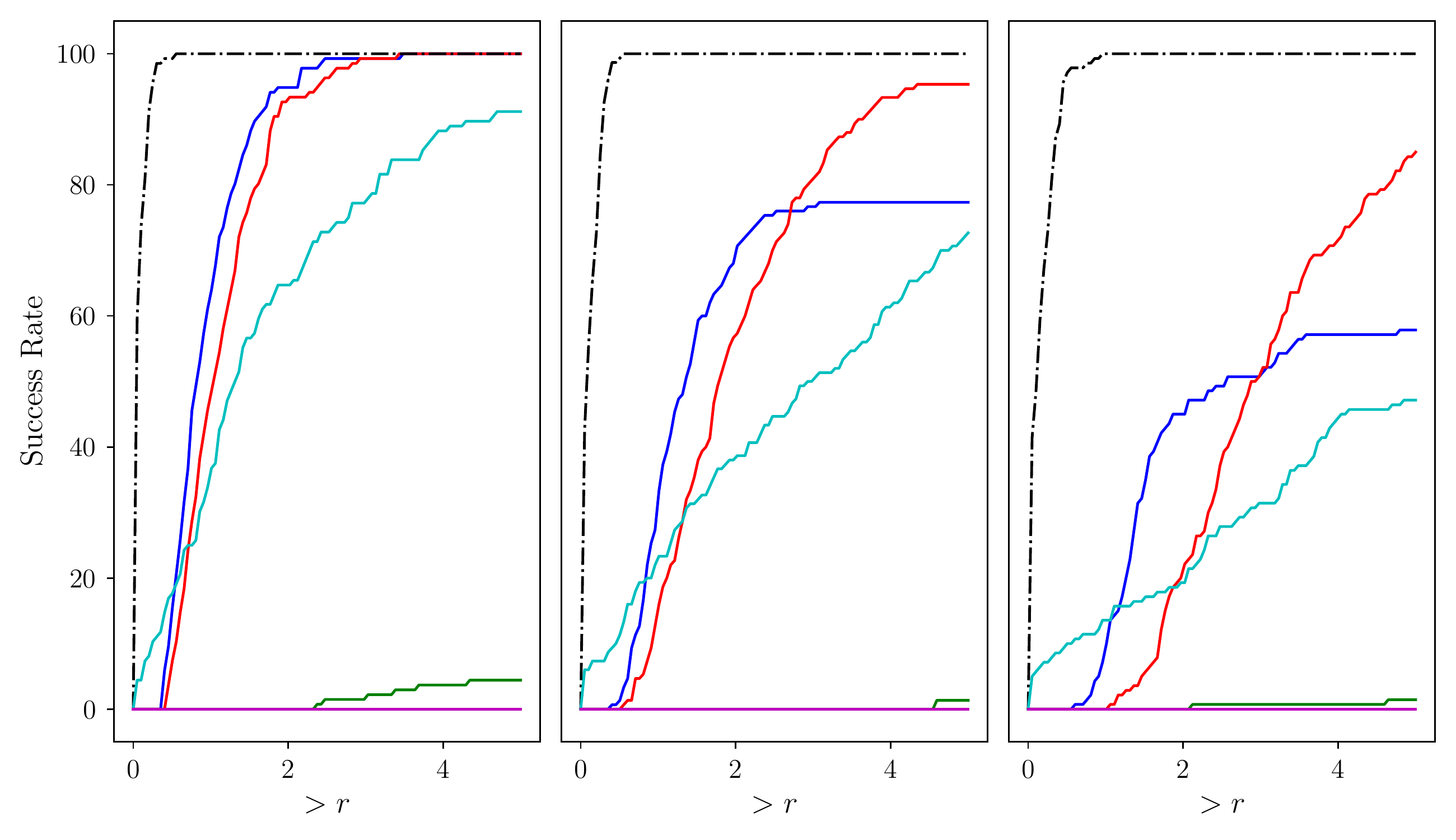} 
  \caption{Attack and certification performance for a ResNet-$110$ model for CIFAR-$10$, when trained with MACER, covering our new Certification Aware Attack (blue), PGD (red), DeepFool (cyan), Carlini-Wagner (green), and AutoAttack (magenta). Similar to Figure~\ref{fig:consolidated}, an ideal attack will approach the \citet{cohen2019certified} radii suggested by the black dotted lines.}\label{fig:macer}   
\end{figure*}

To test the performance of our new attack framework against models trained with MACER, Table~\ref{tab:macer_times} and Figure~\ref{fig:macer} recreate earlier results from within this work for CIFAR-$10$, subject to the same form of parameter exploration seen within Appendix~\ref{sec:parameter_space}. We note that these calculations were performed with a ResNet-$110$ architecture, rather than the ResNet-$18$ architecture employed within the previous sections. While the broad qualitative feature of the success rates, best proportions, and median certifications broadly align with those seen within Table~\ref{tab:main_times}, we note that there is a significant difference in the $\%\text{-C}$ scores, which are a product of the ResNet-$110$ architecture (when trained under MACER) producing certifications that are an order of magnitude smaller than those observed within the main body of this work. That the attack radii are remaining constant while the certification radii decrease, strongly suggests that there would be significant scope for improving the performance of these results by varying the range of the parameter  space exploration. One other notable feature is the improvement of DeepFool for MACER trained models, relative to the performance seen within the main body of this work, which we believe is a consequence of the changes in MACER's model decision space influencing the ability for DeepFool to converge upon successful evasion attacks.

\begin{table*}
  \caption{CIFAR-$10$ attack performance across $\sigma$ for a ResNet-$110$ architecture trained with MACER. Table features follow Table~\ref{tab:macer_times}
  } 
  \label{tab:macer_times}
  \centering
  \begin{tabular}{llrrS[table-format=2.2]rS[table-format=2.2]}
    \toprule
$\sigma$ & Type & Suc.$\uparrow$ & Best 
 $\uparrow$ & $r_{50} \downarrow$ & \%\text{-C}$\downarrow$ & $\text{Time}\downarrow$ \\
\cmidrule(r){1-2} \cmidrule(r){3-7}
$0.25$ & $\text{Ours}^{\star}$ & $100 \%$ & $76 \%$  & 0.83 & $2308$ & 9.66 \\
 & $\text{PGD}^{\star}$ & $100 \%$ & $5 \%$  & 1.03 & $2918$ & 24.47 \\
& $\text{C-W}^{\star} $ & $24 \%$ & $0 \%$  & 9.10 & $39952$ & 24.57 \\
& $\text{DeepF} $ & $100 \%$ & $18 \%$  & 1.32 & $3687$ & 7.01  \\
\cmidrule(r){1-2} \cmidrule(r){3-7}
$0.5$ & $\text{Ours}^{\star}$ & $77 \%$ & $58 \%$  & 1.09 & $2875$ & 12.18 \\
 & $\text{PGD}^{\star}$ & $95 \%$ & $18 \%$  & 1.73 & $2294$ & 24.74 \\
 & $\text{C-W}^{\star}$ & $43 \%$ & $1 \%$  & 11.35 & $19073$ & 24.83 \\
 & $\text{DeepF}$ & $100 \%$ & $23 \%$  & 2.94 & $4377$ & 7.58\\
\cmidrule(r){1-2} \cmidrule(r){3-7}
$1.0$ & $\text{Ours}^{\star}$ & $59 \%$ & $43 \%$  & 1.38 & $12654$ & 14.06\\
 & $\text{PGD}^{\star}$ & $98 \%$ & $39 \%$  & 2.86 & $3201$ & 24.52\\
 & $\text{C-W}^{\star}$ & $9 \%$ & $0 \%$  & 9.63 & $20597$ & 24.60\\
 & $\text{DeepF}$ & $100 \%$ & $19 \%$  & 5.29 & $5670$ & 7.11\\
    \bottomrule
  \end{tabular}
\end{table*}

\section{Exemplar Attacks}\label{sec:examplar}

\begin{figure*}[!ht]
    \centering
    \includegraphics[width=.275\linewidth]{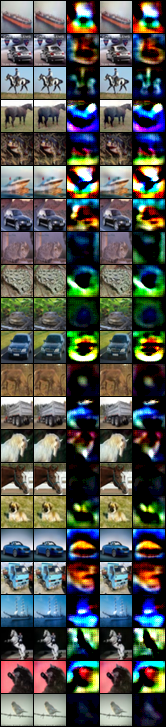} 
  \caption{Illustrative examples of attack performance. Each column (from left to right) represents: the original image; the image under our attack; the adversarial perturbation associated with our attack; the image under PGD; the adversarial perturbation associated with PGD. The adversarial perturbations have been multiplied by $25$ for visual clarity. 
  }\label{fig:illustrative_examples}   
\end{figure*}

The size of the associated adversarial perturbations has been established as a proxy of the risk of an adversarial attack evading human-or-machine-scrutiny~\cite{gilmer2018motivating}. While considering metrics of performance are a more reliable measure of this adversarial risk, for completeness in Figure~\ref{fig:illustrative_examples} provides a visual exemplar of the performance of both our attack and PGD. As both attacks share similar methodological features, the adversarial perturbations share similar semantic features, however our attack consistently requires smaller adversarial perturbations in order to trick the classifier---which in turn would have a higher probability of potentially evading any detection framework. 

\end{document}